\documentclass[runningheads]{llncs}
\usepackage{graphicx}
\usepackage{xcolor}
\usepackage{hhline}
\usepackage{scalerel,xparse}
\usepackage{hyperref}

\usepackage{wrapfig}
\usepackage{subcaption}
\usepackage{arydshln}
\usepackage{marvosym}
\hypersetup{colorlinks=true}

\newcommand{\thickhline}{%
    \noalign {\ifnum 0=`}\fi \hrule height 1pt
    \futurelet \reserved@a \@xhline
}

\NewDocumentCommand\emojimaison{}{\scaleto{\includegraphics{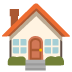}}{0.3cm}}

\NewDocumentCommand\emojihopital{}{\scaleto{\includegraphics{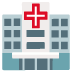}}{0.3cm}}

\NewDocumentCommand\emojifirstname{}{\scaleto{\includegraphics{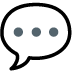}}{0.3cm}}

\NewDocumentCommand\emojimois{}{\scaleto{\includegraphics{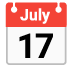}}{0.3cm}}

\NewDocumentCommand\emojiadmin{}{\scaleto{\includegraphics{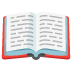}}{0.3cm}}

\NewDocumentCommand\emojiannee{}{\scaleto{\includegraphics{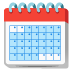}}{0.3cm}}

\NewDocumentCommand\emojiname{}{\scaleto{\includegraphics{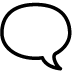}}{0.3cm}}

\NewDocumentCommand\emojicarte{}{\scaleto{\includegraphics{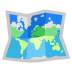}}{0.3cm}}

\NewDocumentCommand\emojiroad{}{\scaleto{\includegraphics{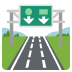}}{0.3cm}}

\NewDocumentCommand\emojisoleil{}{\scaleto{\includegraphics{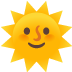}}{0.3cm}}

\NewDocumentCommand\emojinumber{}{\scaleto{\includegraphics{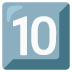}}{0.3cm}}

\NewDocumentCommand\emojicity{}{\scaleto{\includegraphics{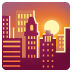}}{0.3cm}}

\NewDocumentCommand\emojihusband{}{\scaleto{\includegraphics{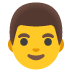}}{0.3cm}}

\NewDocumentCommand\emojibride{}{\scaleto{\includegraphics{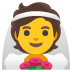}}{0.3cm}}

\NewDocumentCommand\emojifather{}{\scaleto{\includegraphics{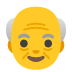}}{0.3cm}}

\NewDocumentCommand\emojimother{}{\scaleto{\includegraphics{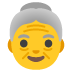}}{0.3cm}}

\NewDocumentCommand\emojiexhusband{}{\scaleto{\includegraphics{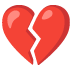}}{0.3cm}}

\NewDocumentCommand\emojistreetname{}{\scaleto{\includegraphics{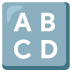}}{0.3cm}}

\NewDocumentCommand\emojimetier{}{\scaleto{\includegraphics{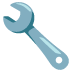}}{0.3cm}}

\NewDocumentCommand\emojihorloge{}{\scaleto{\includegraphics{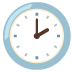}}{0.3cm}}

\NewDocumentCommand\emojitemoin{}{\scaleto{\includegraphics{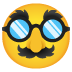}}{0.3cm}}

\NewDocumentCommand\emojireveil{}{\scaleto{\includegraphics{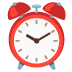}}{0.3cm}}

\NewDocumentCommand\emojisablier{}{\scaleto{\includegraphics{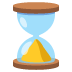}}{0.3cm}}

\begin{document}
\title{End-to-end information extraction in handwritten documents: Understanding Paris marriage records from 1880 to 1940}

\titlerunning{End-to-end information extraction in handwritten documents}
\author{Thomas CONSTUM\inst{1}(\Letter)\orcidID{0009-0006-6623-9366} \and
Lucas PREEL\inst{1}\orcidID{0009-0008-8196-5491} \and
Théo LARCHER\inst{1}\thanks{Théo LARCHER is now 
a research engineer
at 
LIRMM, Montpellier University.} \and
Thierry PAQUET\inst{1}\orcidID{0000-0002-2044-7542} \and
Pierrick TRANOUEZ\inst{1}\orcidID{0000-0002-1962-0782} \and
Sandra BREE\inst{2}\orcidID{0000-0002-2802-5563}
}
\authorrunning{T. CONSTUM et al.}
\institute{LITIS EA4108, University of Rouen Normandy, France
\email{\{ thomas.constum1 , lucas.preel, theo.larcher, thierry.paquet, pierrick.tranouez \}@univ-rouen.fr}\\
\and LARHRA, UMR 5190, CNRS, France \email{sandra.bree@msh-lse.fr}
}

\maketitle              %
\begin{abstract}
The EXO-POPP project aims to establish a comprehensive database comprising 300,000 marriage records from Paris and its suburbs, spanning the years 1880 to 1940, which are preserved in over 130,000 scans of double pages.
Each marriage record may encompass up to 118 distinct types of information that require extraction from plain text.
In this paper, we introduce the M-POPP dataset, a subset of the M-POPP database with annotations for full-page text recognition and information extraction in both handwritten and printed documents, and which is now publicly available.\footnote{\url{https://zenodo.org/record/10980636}}
We present a fully end-to-end architecture adapted from the DAN, designed to perform both handwritten text recognition and information extraction directly from page images without the need for explicit segmentation.
We showcase the information extraction capabilities of this architecture by achieving a new state of the art for full-page Information Extraction on Esposalles and we use this architecture as a baseline for the M-POPP dataset.
We also assess and compare how different encoding strategies for named entities in the text affect the performance of jointly recognizing handwritten text and extracting information, from full pages.


\end{abstract}

\section{Introduction}

In the digital technology era, numerous handwritten collections holding precious information for historians continue to go untapped due to the prohibitive costs and time required for manual analysis.
The EXO-POPP project aims to build a vast database of individuals (300,000 marriage records) from the handwritten marriage records of Paris from 1880 to 1940 that are stored in more than 130,000 scans of double pages. This project will complement the information collected by the POPP project \cite{constum_recognition_2022}, which focused on the Paris censuses between the two World Wars.
Indeed, marriage records contain a large amount of personal information about family histories (bride, groom, parents, and witnesses), such as occupations, date and place of birth, or place of residence
that would allow to analyze the social development of cities throughout history, once made accessible after a digitization and extraction process. 
Marriage records are paragraphs of text (typewritten or handwritten) written in registers one after the other. Accessing each individual information requires locating, transcribing, and tagging the textual units (words) that pertain to each record. These pieces of information are often referred to as named entities, but the information we are looking for extends beyond simple named entities as we wish to detect the relations between the entities as well. Therefore, in this article, we refer to \textit{information extraction (IE)} to name the task at hand.

Traditionally, IE in digitized documents can be decomposed into a three-step processing pipeline: document image segmentation, text recognition (TR), and natural language processing %
applied on the transcriptions. 
One of the main issues with such a processing pipeline is error propagation: each error encountered at a given stage has an impact on subsequent stages. Moreover, sequential approaches require annotated data and training for each stage. Modifying the function at one stage may impact subsequent stages and may require retraining them.
Additionally, when used to process large volumes of data,
sequential approaches
generate intermediate files for each stage, which can require a significant amount of storage, especially for segmentation stages. 

Recently, several methods have emerged as good alternatives to perform handwritten text recognition (HTR) directly at paragraph level \cite{yousef_origaminet,coquenet_end} or at the page level \cite{coquenet_dan} without any segmentation step. In this article, we will refer to these methods as segmentation-free methods but they are often referred to as end-to-end methods in the literature.
The drawback of such approaches is their need for more powerful computational resources due to the size of the input images to deal with at the highest resolution. Moreover, they generally assume a reading order of the document that is sometimes difficult to define as in the MAURDOR dataset \cite{maurdor} for instance.

In parallel, advances have been made to perform both HTR and IE end-to-end at line level \cite{carbonell_joint_2018,wigington_multi-label_2019,boros_comparison_2020,tarride_comparative_2022} or at page level using a Feature Pyramid Network to produce word bounding boxes \cite{carbonell_neural_2020}. These methods are referred to as combined approaches or integrated approaches. We call these methods joint HTR and IE. It has been shown in \cite{carbonell_joint_2018} that %
concatenating the line predictions to expand the context
can improve the performance of IE. 
The article \cite{tarride_comparative_2022} illustrates that
integrated approaches can improve the performance of both HTR and IE.
Nonetheless, the use of integrated methods rather than sequential ones for IE is currently an open debate. For instance, it was demonstrated by \cite{tuselmann_are_2021}, that a sequential approach can outperform an integrated approach through the use of
language models, in this case, a RoBERTa model pre-trained for named entity
recognition (NER).

Finally, two recent attempts have been made to combine a segmentation-free architecture with joint HTR and NER 
by using special tokens to encode named entities.
The first approach \cite{rouhou_transformer-based_2022}, employs a Resnet encoder paired with a transformer decoder and evaluates this architecture on paragraph images from the Esposalles dataset\cite{romero_esposalles_2013}. The second approach \cite{tarride_key-value_2023} builds upon a pre-trained DAN, which is fine-tuned for joint HTR and NER on page images from the Esposalles dataset and paragraph images from the IAM database \cite{marti_iam-database}. 

The Esposalles dataset contains handwritten marriage acts written in old Catalan while the IAM dataset is made of English handwritten paragraphs
sourced
from the LOB corpus\footnote{\url{https://en.wikipedia.org/wiki/Lancaster-Oslo-Bergen_Corpus}}. 
To our knowledge, the IAM and Esposalles datasets are the only open-source handwritten datasets available for evaluating NER tasks at the page level. However, these datasets have several limitations. 

The Esposalles dataset is written by only one author and contains a limited number (13) of named entity categories to extract. Since its publication at the ICDAR 2017 Information Extraction in Historical Handwritten Records competition \cite{fornes_icdar2017_2017},
the Deep Neural Networks architectures for IE on handwritten documents have evolved considerably.
As a result, the
architecture introduced in this article achieves a new state-of-the-art of 96.84\% for the IEHHR score on full pages. (More details about this result and the training procedure can be found in subsection \ref{esposalles}.)
This result suggests that the scientific challenges posed by this dataset are now resolved. 

The IAM dataset, on the other hand, was originally intended only for HTR. This dataset was later annotated with named entities by \cite{tuselmann_are_2021} following the OntoNotes \cite{pradhan_towards_2013} dataset named entity ontology.
Its
weakness
is that it has very sparse named entity annotation: some entire pages contain no named entities, and some types of named entities are very rare.
The IAM dataset allows for evaluating the named entity recognition task on natural language and is not representative of the information extraction tasks found in the field of historical document processing.
Furthermore, these two datasets do not allow a fair segmentation evaluation. Indeed, the IAM dataset contains only a single paragraph per page, and the Esposalles dataset does not contain annotations of the elements located in the margin of the pages.

In this paper, we introduce the M-POPP benchmark dataset (Marriage records of the POPulation of Paris), a new dataset for IE at the page level including both handwritten and typewritten documents. 
The handwritten part includes over a hundred different handwriting styles. Each record can contain up to 118 %
different types of categories to be extracted. The documents are dense in information, as each marriage record contains an average of 48 and 60 different pieces of information for the handwritten and typewritten parts, respectively. Each page has a complex layout with one or several paragraphs per page and two types of text blocks located in the margin whose position can vary greatly from one image to another as shown in Figure \ref{fig:double-page} and Figure \ref{fig:double-page-tapuscrit}. 

In summary, the contributions of our paper are as follows:
\begin{itemize}
    \item We introduce M-POPP, a new dataset for HTR and IE on full pages of historical documents and we make it  publicly available.%
    \item We adapt the DAN architecture for joint HTR and IE on complete documents and achieve a new state of the art on the Esposalles dataset while establishing a strong baseline on the M-POPP dataset.
    \item We compare different methods for named entities encoding and show that these can have a significant impact on information extraction performance.
\end{itemize}

This paper is organized as follows.
Section 2 focuses on presenting the M-POPP corpus and the M-POPP dataset.
Section 3 presents our architecture and evaluation results on the M-POPP dataset.
Section 4 is devoted to the preprocessing stage.

\section{Corpus and ground-truthed datasets}
\subsection{Presentation of the M-POPP corpus}
\label{sec:dataset}
The M-POPP corpus (which stands for Marriage records of the POPulation of Paris) is the corpus on which the EXO-POPP project focuses. 
This corpus was built by gathering the marriage records of Paris and its suburb regions (\textit{Hauts-de-Seine}, \textit{Seine-Saint-Denis}, \textit{Val-de-Marne}).    

\begin{figure}[h]
    \centering
    \includegraphics[width=0.85\textwidth]{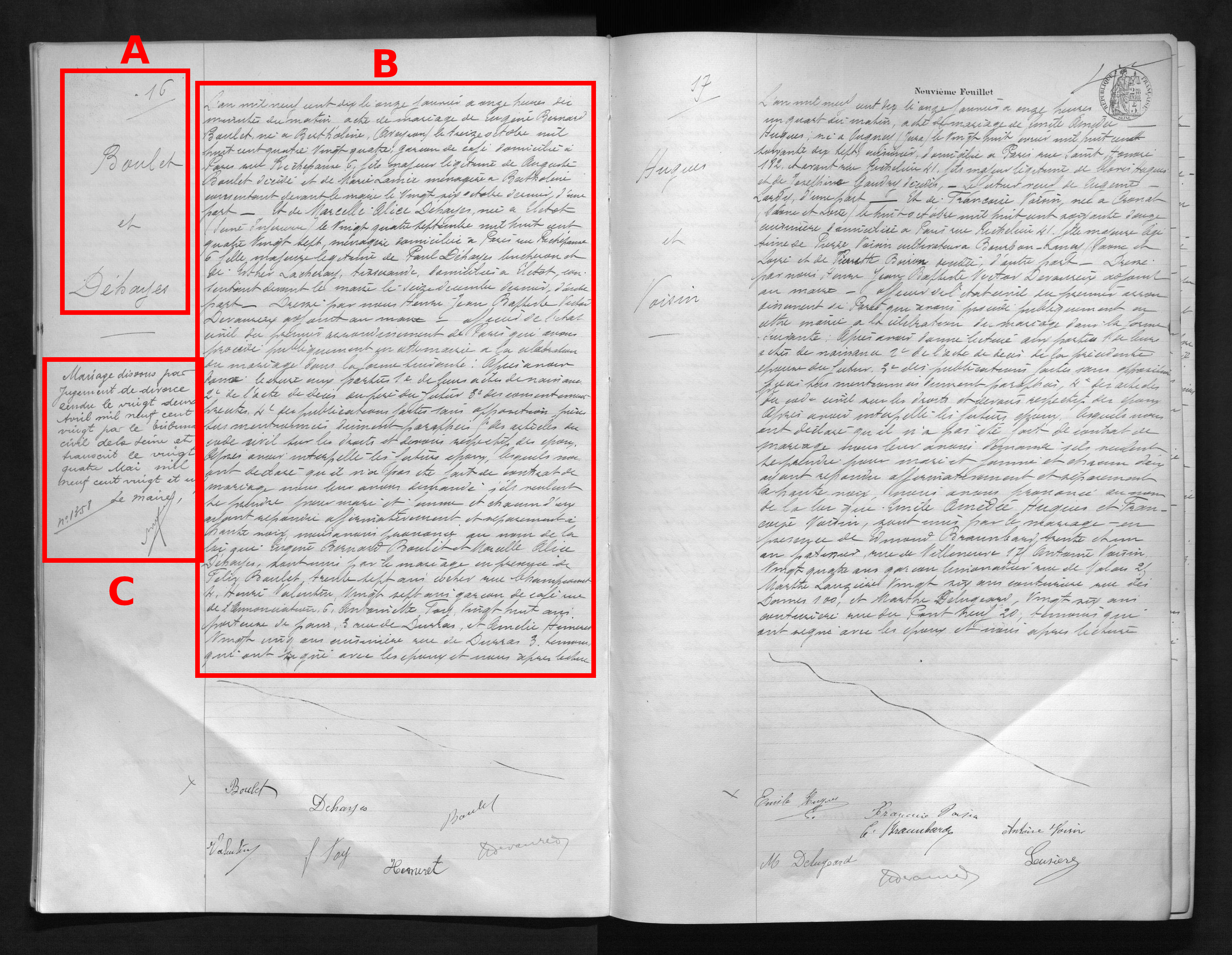}
    \caption{
        An example of a double page containing two handwritten marriage records including one with margin notes (Paris, first \textit{arrondissement}, 1880). 
        The red rectangles give an example of each type of text block (A, B, and C).
    }
    \label{fig:double-page}
\end{figure}

\begin{figure}[]
    \centering
    \includegraphics[width=0.9\textwidth]{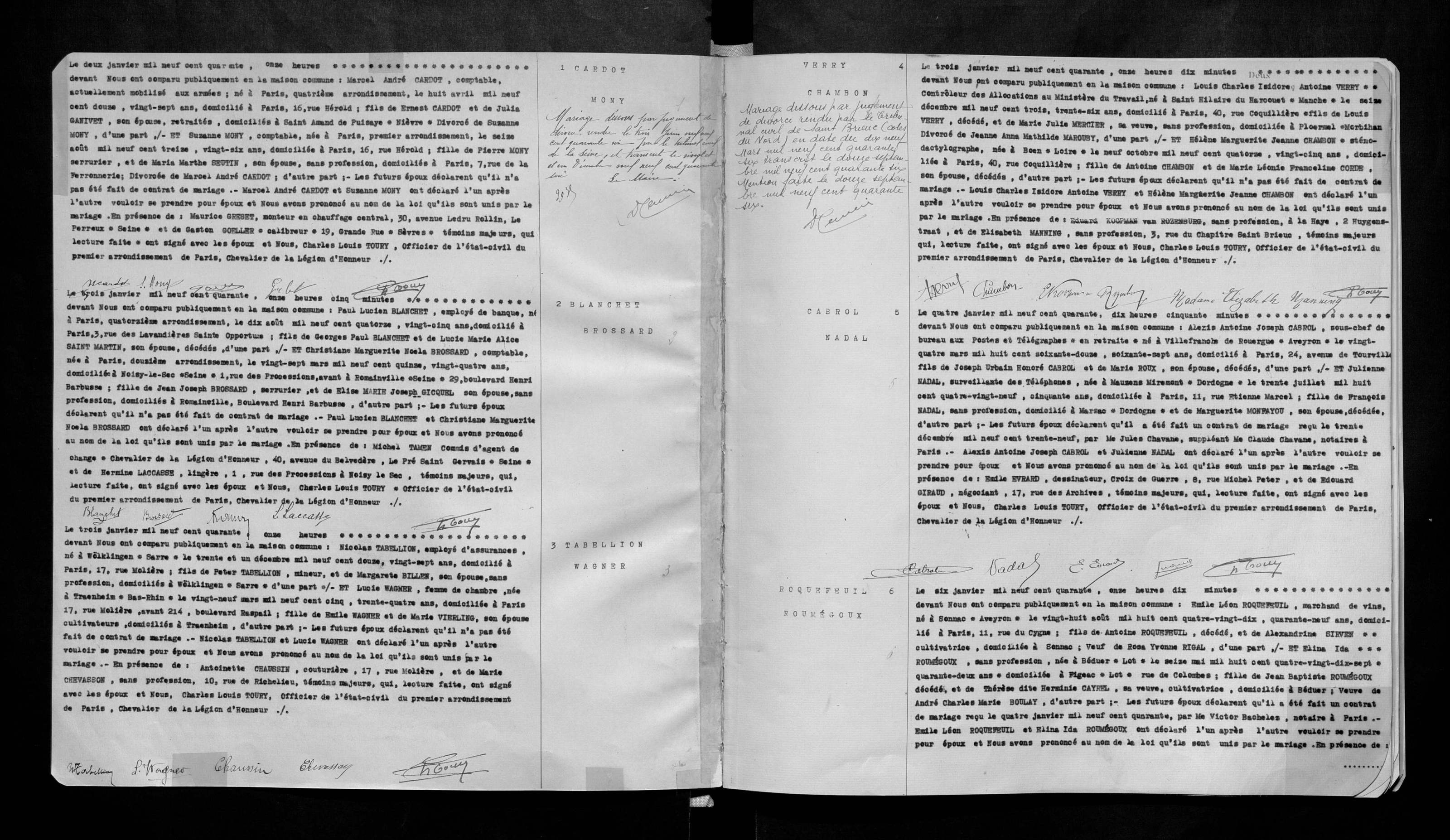}
    \caption{An example of a double page containing six printed marriage records including two with margin notes (Paris, first \textit{arrondissement}, 1940).}
    \label{fig:double-page-tapuscrit}
\end{figure}

The raw dataset stored by the different archives consists of a set of double pages (see Figure \ref{fig:double-page} and \ref{fig:double-page-tapuscrit}) made of every page of the marriage record booklets. %
To reduce the total number of pages to be processed, the dataset was sampled every 10 years. Thus, we focus on documents from the years 1880, 1890, 1900, 1910, 
1920, 1930 and 1940. In total, the sampled corpus contains approximately 300,000 marriage records and 130,000 double pages. 
It contains 12\% of typewritten documents that come mostly from the years 1930 and 1940 of Paris. The other typewritten documents come from the
\textit{Hauts-de-Seine}
region for the year 1930. It should be noted that even in typewritten marriage records, some handwritten information occur, especially concerning the names of the groom and wives, and notes in the margin as seen in Figure \ref{fig:double-page-tapuscrit}.
More details on the geographical distribution of the corpus are detailed in the appendix \ref{appendix-corpus}.

In this study, we focus on the pages of the corpus that contain marriage records.
These marriage acts are made of 3 types of text blocks. Block A is located in the margin and contains the last names of the married couple, possibly with their first names and the date of the marriage. Block B is the body of the text.
Finally, block C is optional and corresponds to marginal notes used in various cases, such as the mention of a divorce or a correction made to the act. Figure \ref{fig:double-page} gives an example of each type of text block.

Block B is the one that contains most of the information to be extracted.\footnote{A small part of the information is located in the margins but is not included in the current version of M-POPP.}
The wording may vary between years and cities, but the key information and their order of appearance in the text are generally stable. In order, there is information about the date of the marriage, information about the husband followed by his parents, and likewise for the wife. Finally, the act ends with some information about the witnesses. 
The marriage act provides geographical information such as a place of birth or residence, temporal information such as a date of birth, as well as various information about individuals such as age or profession.

This corpus is challenging for many reasons. First, it presents a large diversity of layouts. Indeed, each double-page can contain between one and six acts due to the size variability of the booklets: the width of the margins, and the size of the acts on the page can vary greatly. The margin around the booklet in the image may also vary. Moreover, marginal notes pose a real challenge as their position in the image can vary significantly. 

Another difficulty of this corpus lies in the handwritten nature of the documents. The documents have been written by different mayors and deputy mayors of different cities over the years. This means that the corpus contains a lot of different writing styles. 
We estimate the number of different authors to be at least 500 for the complete
corpus due to its geographical and temporal spread.
Moreover, the variation in handwriting size can be very significant. Indeed, marginal annotations tend to be written smaller than the main text, while the names of the husband and wife can sometimes be written in a significantly larger size.

The corpus also presents challenges related to computational resource consumption. Indeed, some images are very large, even at a 150 DPI resolution. For instance, a double-page scan can reach a size of $2200 \times 3000$ pixels at 150 DPI, with a small handwriting size such that further reduction of the image size would destroy the information to be extracted. Such image dimensions result in a significant demand for computational resources, particularly with end-to-end Deep Neural Network architectures.
Additionally, the scanning conditions and quality may vary between regions. For example, some have been scanned in color and some others in grayscale.
Given that most images were in grayscale, we converted all images to this format to help reduce memory usage.

The last challenge is related to information extraction. Indeed, a total of 118 information categories need to be extracted. Moreover, some named entities are very close semantically, such as the \textit{first name of the husband's father} and the \textit{first name of the wife's father}, requiring a deep understanding of the context.

\subsection{Annotation of two datasets}
\label{sec:genericdataset}
To tune the TR model, a training dataset is required. Therefore, we annotated two datasets to represent the corpus.
The first contains handwritten marriage records and the second, typewritten ones. 
As explained in \ref{sec:dataset}, the M-POPP corpus exhibits significant variability despite being composed solely of marriage records.
We diversified the handwritten dataset by years and cities of origin, limiting it to one image per city-year combination to ensure representativeness.
Regarding the typewritten dataset, we only varied the original year of the acts since almost all of them come from Paris.
More details about the datasets can be found in the appendix \ref{appendix-dataset}.

The scans of double pages have been divided into images of single pages. Indeed, as explained earlier, the images in the corpus are very large and have a high density of textual content, which excludes any resizing of the images. Therefore, a division into single pages was necessary to reduce the GPU memory consumption of deep learning trainings as explained in section \ref{preprocessing}.

\subsubsection{Annotation for text recognition}
Since we aim to perform HTR on entire pages, it is necessary to encode the document structure in the annotation. For this purpose, we employ the procedure applied in \cite{coquenet_dan}, which involves adding opening and closing tags to the character set for each text block we want to recognize. In total, we define four types of text blocks, including the three types mentioned in subsection \ref{sec:dataset} and a fourth type, block D, encoding marriage records. More precisely, block D corresponds to a set containing a block A and a block B, optionally with one or more blocks C. An example of annotation for a complete page can be found in the appendix \ref{appendix:exemple-annotation}.

\subsubsection{Annotation for information extraction} \label{ne-annotation}

The annotation of a named entity dataset can be very time-consuming, often much more than that to provide transcription annotations. The dataset contains 118 information categories, annotating each of these categories separately would not be feasible within a reasonable time frame. To simplify the annotation process, the types of named entities have been broken down into nested sub-elements. For example, the named entity type \textit{first name of the husband's father} becomes the combination of the nested types \textit{first name}, \textit{father}, and \textit{husband}. Thus, we broke down the named entities into sub-elements pertaining to 4 hierarchical levels, which reduces the total number of %
categories to 23 instead of 118. Notice that level 1, 2, and 3 categories do not encode named entities but rather the relations that may occur between some lower level categories for example: (day, birth, husband) encodes the fact that the annotated piece of text is the date of birth of the husband. Table \ref{tab:ner-levels} details the hierarchy of sub-elements constituting the named entities along with their naming convention. We also include the corresponding emojis that we use as special tokens to represent each tag in the ground truth of the TR+IE model. More details about this representation can be found in subsection \ref{ne-encoding}.

\begin{table}[]
\centering
\caption{Details of the hierarchical breakdown of named entities. Each tag is placed in the corresponding hierarchical level and associated with the emoji representing it.}
\begin{tabular}{|c|cccc|}
\hline
Level & Tags &  &  &    \\ \hline
1 & Administrative \emojiadmin{}{} & Husband \emojihusband{}{} & Wife \emojibride{}{} & Witness \emojitemoin{}{}  \\ \hline
2 & Father \emojifather{}{} & Mother \emojimother{}{} & Ex-husband \emojiexhusband{}{}%
&   \\ \hline
3 & Birth \emojihopital{}{} & Residence \emojimaison{}{} &  &   \\ \hline
4 &  First name \emojifirstname{}{} & Family name \emojiname{}{} & Age \emojisablier{}{} & Occupation \emojimetier{}{} \\ %
  & Street number \emojinumber{}{} & Street type \emojiroad{}{} & Street name \emojistreetname{}{} & City \emojicity{}{}\\ %
  & \textit{Département} \emojicarte{}{} & Country \emojicarte{}{} & Day \emojisoleil{}{} & Month \emojimois{}{} \\ %
 &  Year \emojiannee{}{} & Hour \emojireveil{}{} & Minute \emojihorloge{}{} & \\ \hline
\end{tabular}%
\label{tab:ner-levels}
\end{table}

\begin{figure}[h!]
    \begin{subfigure}[]{.96\textwidth}%
             \centering
             \includegraphics[width=0.88\linewidth]{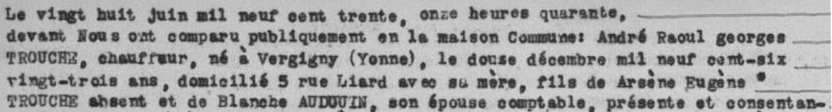}
             \caption{}
             \label{fig:annotation-ner}
    \end{subfigure}
    \begin{subfigure}[]{.98\textwidth}%
             \centering
                \includegraphics[width=0.85\linewidth]{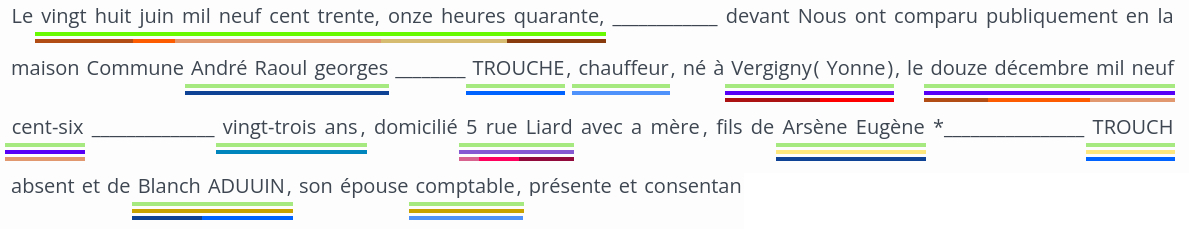}
                
                \includegraphics[width=0.93\linewidth]{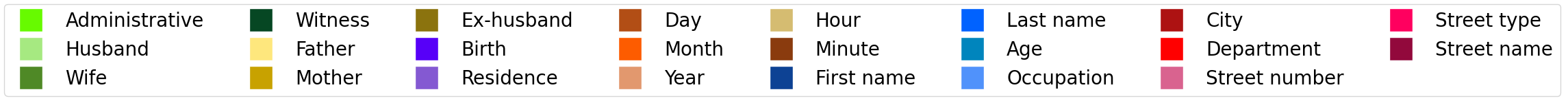}  
            \caption{}
             \label{fig:annotation-ner-pivan}
    \end{subfigure}%
    \caption{(a) Sample from a marriage record from Paris, 1910.
    (b) View of the corresponding annotation for IE in Pivan.
    }
    \label{fig:annotation-et-image-ner}
\end{figure}

The annotation was carried out on the web platform Pivan 2.0 \cite{constum_pivan:_2023}. This version includes a dedicated tool perfectly suited for the annotation of nested named entities. The combination of multiple sub-elements of named entities is represented by the stacking of colored lines under the words, allowing for a clear representation of the annotation.
The annotation is performed by highlighting the relevant words after selecting one or several tags to apply. Figure \ref{fig:annotation-et-image-ner} contains an extract from a marriage record along with the corresponding named entity annotation visualized on Pivan. The number of occurrences of the 
sub-elements
of named entities for both annotated datasets can be found in the appendix \ref{appendix-named-entities}.

\section{Joint Handwriting Recognition and Information Extraction with the DAN}
\subsection{Training procedure and architecture modifications}\label{htr}
We use the DAN \cite{coquenet_dan} as the starting point for our architecture as it is currently the reference architecture to perform handwriting recognition on full pages without any segmentation stage. Let us recall that the DAN architecture is composed of a fully convolutional encoder (FCN) followed by a transformer decoder \cite{vaswani_attention_2023} and can perform both layout analysis and TR end-to-end thanks to the use of specific layout symbols.

Similarly, as DAN deals with extra layout symbols, we expect it to manage additional named entity symbols as well, thus making the system process end-to-end from image to named entities.  
Training a DAN architecture relies on curriculum learning \cite{hacohen_power_2019}, a procedure that starts training the architecture with the 
simplest
examples (single text lines) before providing more complex examples (paragraphs) and finally full document images. As this procedure requires a lot of training data, a synthetic data generator was introduced to start training the system with synthetic printed document images. At this stage, some cursive fonts allow to mimic the cursive style of real handwriting somehow. 

As outlined in the introduction, an adaptation of the DAN has been previously proposed in \cite{tarride_key-value_2023} for jointly performing TR and IE. 
For this purpose, the authors start from a DAN model trained to perform HTR on RIMES \cite{grosicki_results_2009} and apply transfer learning on a new model which is this time trained to perform both TR and IE using special tokens to encode the named entities.
This approach has the advantage of being straightforward and seems sufficient for a simple dataset such as Esposalles. 
In our case, simply applying transfer learning from a DAN trained on another dataset did not work, and we therefore had to make several modifications to the architecture of the DAN and its training process.

First, the curriculum procedure was augmented with one additional step devoted to IE. This means that the system is first trained end-to-end to recognize full document images as the standard DAN, and then it is finally trained to learn to generate the named entity's extra symbols. This process is described in the next paragraph devoted to IE. 
The second modification concerns the synthetic document generator which was adapted to mimic the layout encountered on the M-POPP dataset. To do so, the READ 2016 single-page generator (included in the original code of the DAN\footnote{\url{https://github.com/FactoDeepLearning/DAN}}) was modified, as it generates documents with a layout close to that of the M-POPP corpus. We added several features in the synthetic data generator such as indents that can appear at the beginning of paragraphs or end of lines that can be shorter than other lines.

The third adaptation of the original DAN generator concerns the language. Indeed, the textual content of marriage records is so regular, with very similar sentences such as "ont comparu devant nous en la maison commune" occurring on almost every record, that the transformer decoder would over-fit the language if trained on the ground truth text of the marriage records. Therefore, the synthetic data generator was modified to use sentences from the 1M sentences French Wikipedia corpus\footnote{\url{https://wortschatz.uni-leipzig.de/en/download/French}} compiled by the University of Leipzig. We are also forcing some documents to include similar sentences at several locations in the text as we found that it was stabilizing the attention of the model.

The fourth modification we made concerns the adaptation of the DAN to the resolution of the images. One of the challenges of the M-POPP dataset is to deal with characters of different sizes, which can sometimes be very small even when the image is at its original full size. Recall that the original DAN architecture was designed and tested on the RIMES \cite{grosicki_results_2009}, IAM \cite{marti_iam-database}, and READ 2016 \cite{sanchez_icfhr2016_2016} datasets. We observed the resolution to which the DAN was tuned on these three datasets was too low for the M-POPP dataset where the attention was sometimes spanning over two lines simultaneously. That is why we modified the architecture of the DAN to increase the granularity of attention along the vertical axis. To achieve this, we reduced the stride from 2 to 1 for the block 5 of the encoder. As a result, the sampling rate of the attention was doubled, which prevented sampling two lines of text in the same attention window. Of course, this is made at the expense of increasing the decoding process somehow, but it is a necessity.

\subsubsection{Named entities encoding}\label{ne-encoding}
Once the model is trained for HTR we can adapt it for end-to-end joint HTR and IE.
As explained in \ref{htr}, we encode the named entities in the text with special tags to adapt the DAN for IE.
This approach was proposed for the first time in \cite{carbonell_joint_2018} where several encoding methods have been experimented on the Esposalles dataset. The authors show that using a single tag before each word is better than using opening and closing tags. 
They also demonstrate that encoding in a single tag the category and the person labels was more efficient than using two different tags.
As \cite{carbonell_joint_2018} was using a CRNN-CTC architecture at line images from Esposalles, we believe it is relevant to assert if the observations from the authors generalize to our case at page level with the DAN end-to-end architecture. In the Esposalles dataset, named entities can be described at two levels of information: the category (Example: \textit{first name}) and the person related to the information (Example: \textit{the husband}). As explained in subsection \ref{ne-annotation}, a similar hierarchical breakdown can be made for the information to be extracted in M-POPP but with a higher level of complexity. Indeed, the number of hierarchical levels into which information is divided can reach 5. This is the case, for example, for the information \textit{City of residence of the husband's parents}, which we divide into \textit{City}, \textit{Residence}, \textit{Father}, \textit{Mother}, and \textit{Husband}.

In this paper, we chose to represent these hierarchical elements with emojis. For instance, the information \textit{first name} is represented by the emoji \emojifirstname. The meaning of each emoji can be found in Table \ref{tab:ner-levels}.
To determine the best way to encode named entities in the ground truth, we compare 5 types of encoding.
To illustrate these encodings, 
let's take for instance \textit{Louis Alexandre MOUDEL} that we define as the father of the bride, where \textit{Louis Alexandre} are his two first names, and \textit{Moudel} is his last name.

\textit{1) Single separate tags before each word:}
In this approach, each level of information is indicated by a dedicated tag, and the tags are placed before the word they encode information for. With this encoding, the ground truth for the example would be:\label{enc-before}

\emojifirstname\emojifather\emojibride Louis \hspace{0.2cm} 
\emojifirstname\emojifather\emojibride Alexandre\hspace{0.2cm} 
\emojiname\emojifather\emojibride MOUDEL\\

\textit{2) Single separate tags after each word:}
Similar to the previous approach, except here the tags are placed after the word.
With this encoding the previous example becomes:\label{enc-after}

Louis\emojibride\emojifather\emojifirstname
\hspace{0.2cm} 
Alexandre\emojibride\emojifather\emojifirstname\hspace{0.2cm} 
MOUDEL\emojibride\emojifather\emojiname\\

\textit{3) Open \& close separate tags:}
Here, each word presenting information to be extracted is surrounded by one or more opening and closing tags, where each tag encodes a level of information. So the example would be as:\label{enc-begend}

\textless\emojibride\textgreater\hspace{0.05cm} \textless\emojifather\textgreater\hspace{0.05cm} \textless\emojifirstname\textgreater Louis \textless\textbackslash\emojifirstname\textgreater\hspace{0.05cm} \textless\textbackslash\emojifather\textgreater\hspace{0.05cm} \textless\textbackslash\emojibride\textgreater

\textless\emojibride\textgreater\hspace{0.05cm} \textless\emojifather\textgreater\hspace{0.05cm} \textless\emojifirstname\textgreater Alexandre \textless\textbackslash\emojifirstname\textgreater\hspace{0.05cm} \textless\textbackslash\emojifather\textgreater\hspace{0.05cm} \textless\textbackslash\emojibride\textgreater

\textless\emojibride\textgreater\hspace{0.05cm} \textless\emojifather\textgreater\hspace{0.05cm} \textless\emojiname\textgreater MOUDEL \textless\textbackslash\emojiname\textgreater\hspace{0.05cm} \textless\textbackslash\emojifather\textgreater\hspace{0.05cm} \textless\textbackslash\emojibride\textgreater\\

\textit{4) Nested open \& close separate tags:}
Similar to the previous approach, but this time a tag is closed only when the encoded information is no longer the same for that level of information. We can see in the example below that the tags for \textit{wife} and \textit{father} are only used twice.\label{enc-begend-nested}

\textless\emojibride\textgreater\hspace{0.05cm} \textless\emojifather\textgreater\hspace{0.05cm} \textless\emojifirstname\textgreater Louis Alexandre \textless\textbackslash\emojifirstname\textgreater

\textless\emojiname\textgreater MOUDEL \textless\textbackslash\emojiname\textgreater\hspace{0.05cm} \textless\textbackslash\emojifather\textgreater\hspace{0.05cm} \textless\textbackslash\emojibride\textgreater\\

\textit{5) Single combined tags after each word:}
In the last approach, one tag encodes all the hierarchical levels constituting information. The tags are located after the word they encode information for. \label{enc-no-hierarchy}

Louis\textless wife\_father\_first\_name\textgreater \hspace{0.05cm} Alexandre\textless wife\_father\_first\_name\textgreater

MOUDEL\textless wife\_father\_family\_name\textgreater

\subsection{Experiments and results}
\subsubsection{Evaluation metrics}
To evaluate the performance of our method for TR we use the CER and the WER.
For IE evaluation, we use the Nerval library\cite{nerval2021} which computes the
F1 metric.
Nerval computes this metric by aligning each prediction with the corresponding ground truth at the character level and by using a 30\% CER threshold to accept the ground truth and the prediction as a positive match.
To allow a fair comparison, every named entity encoding (prediction and ground truth) has been formatted to the same Beginning Inside Outside (BIO) encoding before processing them by Nerval.

\subsubsection{Preliminary experiment}\label{esposalles}
Before training our model on the M-POPP dataset, we conducted evaluations using the Esposalles dataset 
to highlight the robust capabilities of the chosen architecture.
To encode the named entities in the ground truth we use special tokens placed after each named entity to extract. We use single separate tags meaning that for each named entity type, there is a tag encoding the person and a tag encoding the category. This encoding is similar to the encoding type 2 described in subsection \ref{ne-encoding}.
Regarding the training procedure, we first train the DAN for HTR before training it for HTR+NER.
To train the model for HTR, we use a synthetic data generator reproducing the type of layout encountered in the Esposalles dataset. The labels used for the text of the synthetic data are drawn from a corpus of 28k sentences from the Catalan Wikipedia.\footnote{\url{https://wortschatz.uni-leipzig.de/en/download/Catalan\#cat_wikipedia_2021}}
We detail in Table \ref{tab:esposalles} the results obtained by our method on the Esposalles dataset, as well as previous state-of-the-art results at line, record, and page levels. The results are expressed on the test set in terms of IEHHR score which was introduced in the ICDAR 2017 competition on Information Extraction in Historical Handwritten Records \cite{fornes_icdar2017_2017}.
We can see that our method achieves a new state-of-the-art on this dataset for information extraction applied to pages. 
It is noteworthy that \cite{tarride_key-value_2023} methodology is akin to ours, utilizing a similar adaptation of the DAN model for IE. We attribute the enhanced performance of our method over \cite{tarride_key-value_2023} to the use of our dedicated synthetic data generator.

\begin{table}[]
    \centering
    \caption{Evaluation results for the Esposalles dataset. Results are given for the IEHHR scores on the test set.}
    \label{tab:esposalles}
    \begin{tabular}{|c|c|c|c|}
        \hline
        Method & Basic (\%) & Complete (\%) & Input type \\ \hline
        Naver Labs \cite{prasad_bench-marking_2018} & 95.46 & 95.03 & Line \\ \hline
        Teklia \cite{tarride_key-value_2023} & 97.03 & 96.93 & Record \\ \hline
        Teklia \cite{tarride_key-value_2023} & 95.45 & 95.04 & Page \\ %
        Ours & \textbf{96.80} & \textbf{96.84} & Page \\ \hline
    \end{tabular}%
\end{table}

\subsubsection{Experiments}
Now that we have validated the effectiveness of our architecture, we can evaluate it on the M-POPP dataset.
For TR, we train our architecture separately on the handwritten dataset and the printed dataset. 
First, for each type of writing, we conduct training where the ground truth includes all three types of text blocks (A, B, C) to demonstrate the layout challenges posed by M-POPP.
Then, for the printed dataset, we remove the text blocks A and C from the ground truth to consider only the main body text since most of the margin blocks are handwritten. This experiment is interesting because even though printed text recognition is no longer a challenge, it allows us to evaluate the impact that TR errors can have on IE by comparing it with the results obtained from handwritten text. This is also the setting used to evaluate IE on printed text.
Regarding the handwritten dataset, we subsequently conducted experiments without block C since this block does not contain any information to extract. 
Moreover, the layout complexity introduced by these blocks could potentially complicate the recognition task, thereby precluding an equitable comparison of TR performance between the HTR and combined HTR+IE scenarios.

\subsubsection{Text recognition performance}
The evaluation results for handwritten text recognition and printed text recognition are detailed in Table \ref{tab:htr-table}.
This table also includes the text recognition results for the TR+IE experiments on the printed and handwritten datasets. It is important to note that CER and WER computations do not include NER tokens. For clarity, we only include the performance obtained with the encoding type 5. The complete TR results for the TR+IE experiments can be found in Table
\ref{tab:ner-table}.%

The results of purely printed text recognition are very good, which can be explained by the low variability of the printed fonts used in the documents. When considering marginal text blocks (blocks A and C) in the printed dataset, we observe a significant degradation of the  performance, due to their greater difficulty in terms of handwriting style and layout.
The CER increases from 0.88\% to 1.54\% when the recognition task is augmented with the Information Extraction (TR+IE). In any case, these results appear very good despite the typewritten texts being not always horizontal, meaning that the layout is well detected by the DAN architecture.    

As expected, the results for HTR are lower than for printed text. The CER (7.42\% for HTR and 6.57\% for HTR+IE) is higher than that achieved by the DAN on other datasets (IAM, RIMES, READ2016), which can be attributed to the significant diversity in handwriting styles of the M-POPP dataset. M-POPP is made of real samples of handwritten texts, whereas IAM and RIMES datasets have been designed to be HTR datasets. Moreover, M-POPP contains very small characters in the margins, which makes recognition more difficult with a 13.75\% CER when including the margin notes.
Contrary to the printed text, we can notice a 0.85\% CER \textit{decrease} for the handwritten text between the pure TR task and the fully integrated TR+IE task.

\begin{table}[]
    \centering
    \caption{Text Recognition (TR) results for different configurations.}
    \label{tab:htr-table}
    \begin{tabular}{|c|c|c|c|c|c|c|c|}
        \hline
        \begin{tabular}[c]{@{}c@{}}Evaluation\\ dataset \end{tabular}& Task & \begin{tabular}[c]{@{}c@{}}With\\block A\end{tabular} & \begin{tabular}[c]{@{}c@{}}With\\block C\end{tabular} &\begin{tabular}[c]{@{}c@{}}CER\\ valid (\%)\end{tabular} & \begin{tabular}[c]{@{}c@{}}WER\\ valid (\%)\end{tabular}& \begin{tabular}[c]{@{}c@{}}CER\\ test (\%)\end{tabular} & \begin{tabular}[c]{@{}c@{}}WER\\ test (\%)\end{tabular} \\ \hline
        Printed & TR & yes & yes & 6.51 & 9.22 & 5.62 & 7.58 \\ %
        Printed & TR & no & no & 0.98 & 3.21 & 0.88 & 3.17 \\ %
        Printed & TR+IE & no & no & 1.53 & 3.9 & 1.54 & 3.55\\ \hline
        Handwritten & TR & yes & yes & 10.21 & 21.61 & 13.75 & 25.94\\ %
        Handwritten & TR & yes & no & 7.1 & 16.84 & 7.42 & 16.29 \\ %
        Handwritten & TR+IE & yes & no & 5.63 & 15.91 & 6.57 & 15.93\\ \hline
    \end{tabular}%
\end{table}

\subsubsection{Information extraction results}\label{ssec:ie-results}
The evaluation results for IE are detailed in Table \ref{tab:ner-table}. For the test sets, the best encoding achieves a 93.04\% F1 score on the typewritten dataset, while this same encoding achieves a 73.51\% F1 score on the handwritten dataset.
From the results of the different encoding methods, we can observe a clear performance gap between
opening and closing tags encoding, and single tag encoding,
which aligns with the findings of \cite{carbonell_joint_2018}. The weakness of opening and closing tags stems from the fact that the omission of a closing tag propagates the assignment of a category till the end of the text, which impacts the metric. This can be dramatic for the M-POPP dataset due to the large size of marriage records.
This performance decrease is all the more important for HTR, where text recognition difficulties increase the risk of forgetting some closing tags.
Regarding the position of the tag relative to the word in question, there is a slight improvement when the tag is put after the word rather than before. This may come from the fact that by predicting the semantic class of a word after the word itself, the DAN can use its self-attention on the word to enhance its tag prediction.
As for the combined encoding compared to the approach with separate tags, we also find the observations made in \cite{carbonell_joint_2018}, with an F1 improvement of 2.19 and 2.67 
for the printed and handwritten datasets respectively between the encoding types 2 and 5.

The comparison of results between the handwritten dataset and the printed dataset allows for evaluating the impact of recognition errors on IE, with a difference of 19.53 F1 points for the test set with encoding type 5. However, it should be noted that other factors could contribute to this difference in results.
For instance, the handwritten dataset exhibits a greater language diversity in the acts, which complicates the model's ability to understand the structure of marriage records. Therefore, expanding the training dataset could enhance the model's performance.
Finally, while there are minor variations in text recognition results among different encoding methods, these discrepancies are likely attributable to stochasticity in training rather than to the encoding methods themselves.

\begin{table}[]
\centering
\caption{Information extraction results for different configurations.}
\label{tab:ner-table}
\begin{tabular}{|c|c|c|c|c|c|}
\hline
Writing type & \begin{tabular}[c]{@{}c@{}}Entity encoding\\ (Encoding type No.)\end{tabular} & \begin{tabular}[c]{@{}c@{}}Hierarchical\\ entities\end{tabular} & \begin{tabular}[c]{@{}c@{}}CER\\ test (\%)\end{tabular}  & \begin{tabular}[c]{@{}c@{}}WER\\ test (\%)\end{tabular} &\begin{tabular}[c]{@{}c@{}}F1\\ test (\%)\end{tabular} \\ \hline
Printed  & before (1)& yes   & 0.94   & 2.81 & 90.21 \\ %
Printed  & after (2)& yes &   1.88 & 3.9 &   90.85 \\ %
Printed  & after and before (3)& yes &   1.94   & 3.56 & 79.63 \\ %
Printed & after and before nested (4)& yes &  1.27 & 3.16   & 75.33 \\ %
Printed  & after (5)& no   & 1.54  & 3.55 & 93.04 \\ \hline%
Handwritten  & before (1)& yes   & 6.85   & 16.14 & 69.81 \\ %
Handwritten  & after (2)& yes &   6.91 & 16.4 &   70.84 \\ %
Handwritten  & after and before (3)& yes &   6.67   & 15.86 & 33.51 \\ %
Handwritten  & after and before nested (4)& yes &  6.7   & 16.03 & 32.5 \\ %
Handwritten  & after (5)& no &  6.57 & 15.93 & 73.51 \\ \hline
\end{tabular}%
\end{table}

\section{Preprocessing}\label{preprocessing}
Even though the DAN can process double-page images, a preprocessing step is necessary to deal with the large size of the M-POPP dataset images.
Indeed, a significant portion of the images from the handwritten dataset is too large to fit into a GPU during DAN training, even with a batch size of 1 and an NVIDIA A100 GPU.
Moreover,
distributing the model on a GPU and the data on another GPU would be a very complex task to implement.
Therefore, we chose to split each double-page image into two single pages, thus doubling the number of images but reducing their size.
It is important to note that this preprocessing step would not be necessary for smaller images or with more GPU memory available. 
To extract single-pages from double-page scans, we used
YOLOv8l \cite{terven_comprehensive_2023} since it is fast and easy to use via Ultralytics\footnote{\url{https://github.com/ultralytics/ultralytics}}, and achieves competitive results on various datasets for object detection on images. 
We manually annotated 90 images (65 images for training, and 25 images for validation) and trained the model to predict the page bounding boxes for 200 epochs.
We obtained a $\mathrm{mAP_{50}}$
\footnote{The $\mathrm{mAP_{50}}$ is the mean Average Precision of detection at an Intersection Over Union (IoU) threshold of 50\%. The $\mathrm{mAP_{50-95}}$ is the  mAP across different IoU thresholds from 50\% to 95\% in increments of 5\%.}
of 1 and a 
$\mathrm{mAP_{50-95}}$
of 0.89 on the validation set. 
An example of pre-preprocessing can be found in Figure \ref{fig:segmentation-example} in the appendix \ref{appendix-preprocessing}.

\section{Conclusion}
In this article, we introduced M-POPP, a new dataset created as part of the EXO-POPP project.
This dataset is made of marriage acts of Paris (1880-1940) containing full-page annotation for HTR and IE and is now publicly available for the community. This corpus encompass up
to 118 distinct types of information that require extraction from plain text.
We presented a fully end-to-end architecture to jointly perform Layout Analysis, Text Recognition, and Information Extraction based on DAN that achieves a new state of the art for NER on Esposalles. With some adaptations of its attention grid, the DAN architecture establishes a strong baseline on the M-POPP dataset.
This study is one of the first attempts towards using end-to-end Information Extraction on full pages of printed and handwritten documents, and the experimentation shows
promising results that demonstrate the interest of this approach. 
This paper also provides experimental results with several encodings of named entities and shows that the best encoding is to combine every hierarchical level of information at once using specialized tags after each word to tag.

In the future, we plan to compare these results with sequential methods using language models such as RoBERTa \cite{liu_roberta:_2019}, or large language models such as LLaMA \cite{touvron_llama:_2023}, for NER. Since the main advantage of sequential methods is to leverage language models, it could also be interesting to integrate them into the end-to-end DAN architecture.

\section{Acknowledgements}
This work was performed using HPC resources from GENCI-IDRIS (Grant 2023-AD011013149R2). This work was financially supported by the Normandy regional council and the ANR EXO-POPP project, grant ANR-ANR-21-CE38-0004 (French Agence Nationale de la Recherche).
\bibliographystyle{splncs04}
\bibliography{bibliography}

\newpage
\section{\appendixname}
\subsubsection{Details about the M-POPP corpus}\label{appendix-corpus}

We provide in Table \ref{tab:info-corpus} the number of images in the corpus according to region of origin and type of writing. It should be noted that most of the images are handwritten and that the majority come from Paris.

\begin{table}[h]
    \centering
    \caption{Number of images in the corpus according to region of origin and type of writing.}
    \begin{tabular}{|c|c|c|c|}
\hline
Region     & Type of writing & \# of double pages & \# of cities/arrondissements \\ \hline
Paris             & Handwritten     & 71.6k              & 20                           \\ \hline
Paris             & Typewritten     & 15k                & 20                           \\ \hline
\textit{Hauts-de-Seine}    & Handwritten     & 18k                & 27                           \\ \hline
\textit{Hauts-de-Seine}    & Typewritten     & 929                & 8                            \\ \hline
\textit{Seine-Saint-Denis} & Handwritten     & 10k                & 24                           \\ \hline
\textit{Val-de-Marne}      & Handwritten     & 11,7k              & 28                           \\ \hline
\end{tabular}
    \label{tab:info-corpus}
\end{table}

\subsubsection{Details about the M-POPP dataset}\label{appendix-dataset}

We provide information on the average characteristics of each record in Table \ref{tab:average-info-acts} and details on the split of the two datasets in Table \ref{tab:info-split-dataset}.

\begin{table}[]
    \centering
    \caption{Average annotation statistics per act for the two M-POPP datasets.}
    \label{tab:average-info-acts}
    \begin{tabular}{|c|c|c|c|}
        \hline
        Dataset & \# of characters & \# of words & \# of named entities \\ \hline
        Handwritten & 1519 & 231 & 48 \\ \hline
        Printed & 1328 & 200 & 60 \\ \hline
    \end{tabular}
\end{table}

\begin{table}[]
    \caption{Images by split for the two M-POPP datasets.}
    \label{tab:info-split-dataset}
    \begin{subtable}{0.45\textwidth}
        \centering
        \caption{Handwritten dataset}
        \begin{tabular}{|c|c|c|c|}
            \hline
             & Train & Validation & Test \\ \hline
            Pages & 250 & 32 & 32 \\ \hline
            Acts & 344 & 51 & 53 \\ \hline
            Named entities & 16727 & 2223 & 2517 \\ \hline
        \end{tabular}%
    \end{subtable}
    \begin{subtable}{0.45\textwidth}
        \centering
        \caption{Printed dataset}
        \begin{tabular}{|c|c|c|c|}
            \hline
            & Train & Validation & Test \\ \hline
            Pages & 116 & 14 & 13 \\ \hline
            Acts & 363 & 43 & 30 \\ \hline
            Named entities & 22036 & 2559 & 2405 \\ \hline
        \end{tabular}%
    \end{subtable}
\end{table}

\subsubsection{Example of page annotation in M-POPP}\label{appendix:exemple-annotation}

We provide an example of HTR annotation for a complete page of the M-POPP dataset. This example illustrates the use of special tags to encode the structure of marriage certificates.

\begin{verbatim}
    <D><A>595 Jegou et Boulin</A>
    <B>Le vingt Février mil neuf cent vingt, seize heures devant
    [...] époux et nous Alphonse Louis Malètre maire-adjoint du
    dix-septième arrondissement de Paris</B>
    <C>680. Mariage dissous par Jugement de divorce rendu le
    [...] Le Maire</C></D> 
    <D><A>787 Delagarde et Meslé</A>
    <B>L'an mil huit cent quatre vingt-dix le quatre novembre
    à dix, [...] le père de l'épouse et Nous, après lecture.</B>
    <C>Approuvé la rature de quinze mots nuls.</C></D>
\end{verbatim}

\subsubsection{Details about the information extraction annotations}\label{appendix-named-entities}

\begin{figure}[h]
    \centering
    \includegraphics[width=\textwidth]{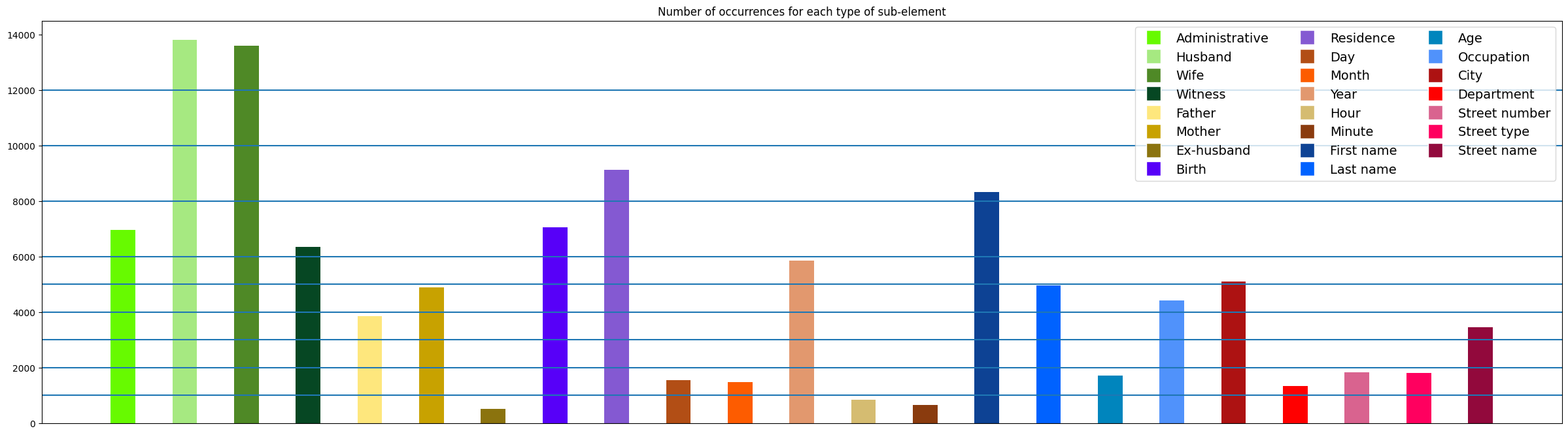}
    \caption{Number of occurrences for each type of named entity sub-element in the printed dataset.}
    \label{fig:nb-occurences}
\end{figure}

\begin{figure}[h!]
    \centering
    \includegraphics[width=\textwidth]{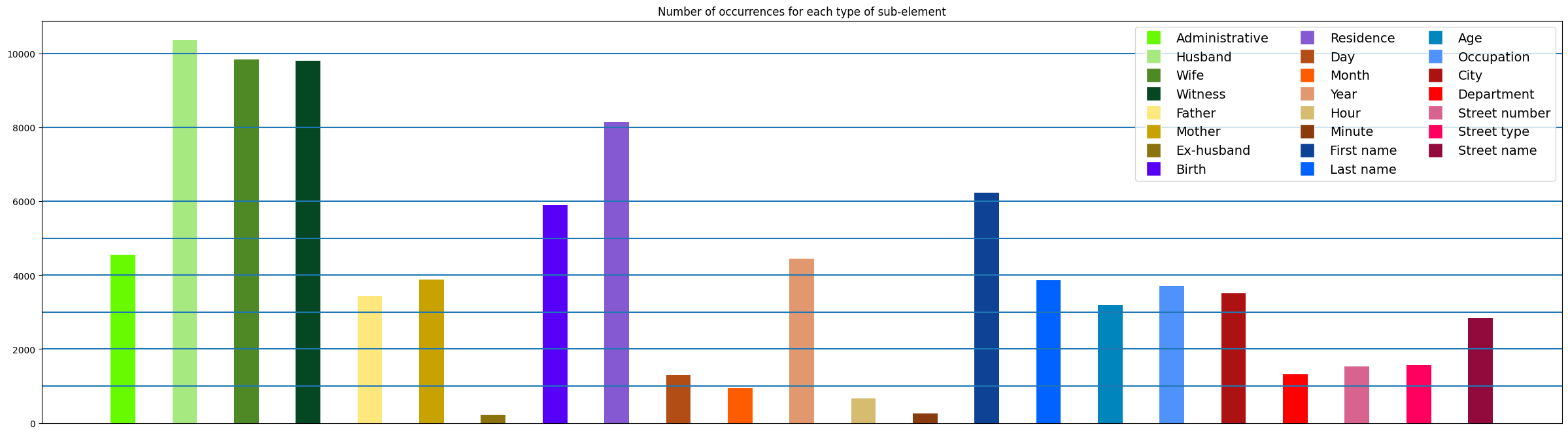}
    \caption{Number of occurrences for each type of named entity sub-element in the handwritten dataset.}
    \label{fig:nb-occurences-htr}
\end{figure}

The number of occurrences for the sub-elements of named entities of both annotated datasets can be found in the diagrams of Figure \ref{fig:nb-occurences} and Figure \ref{fig:nb-occurences-htr}.

\newpage
\subsubsection{Example of preprocessing}\label{appendix-preprocessing}
\begin{figure}[!h]
    \begin{subfigure}[]{.45\textwidth}
             \centering
             \includegraphics[width=0.98\linewidth]{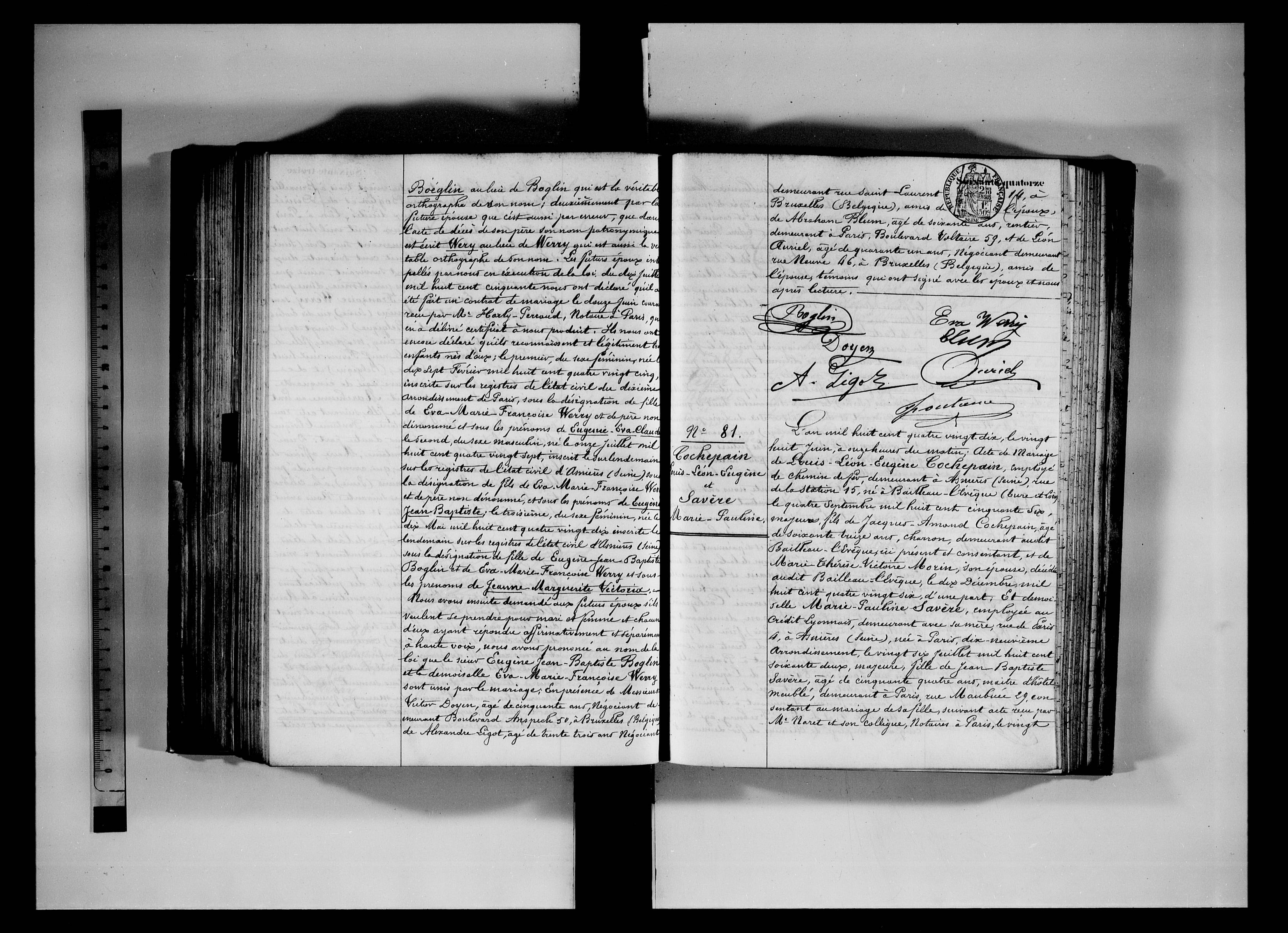}
             \caption{}
             \label{fig:full-double-page}
    \end{subfigure}
    \begin{subfigure}[]{.45\textwidth}
             \centering
                \includegraphics[width=0.98\linewidth]{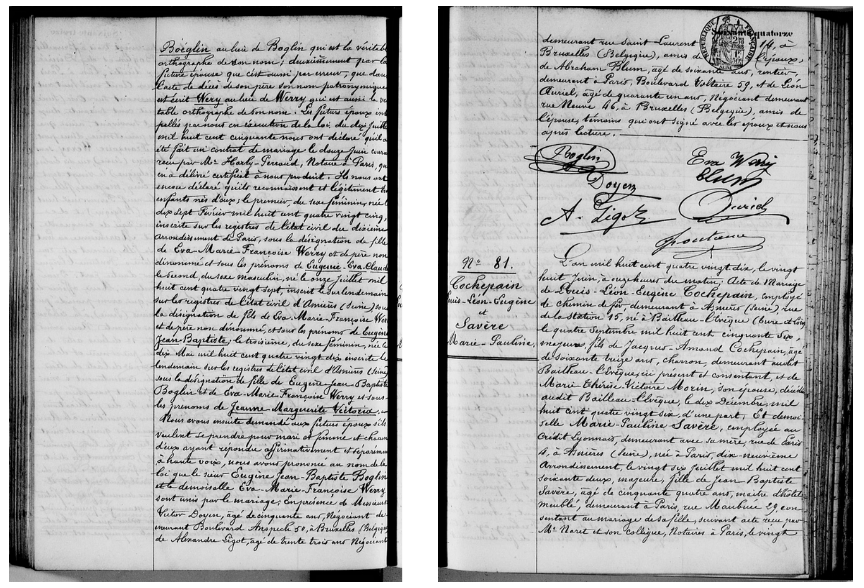}
            \caption{}
             \label{fig:segmented-image}
    \end{subfigure}%
    \caption{(a) Full double-page image
    (b) Preprocessing result
    }
    \label{fig:segmentation-example}
\end{figure}

Figure \ref{fig:segmented-image} shows the segmented single-page images obtained from the double-page image of Figure \ref{fig:full-double-page}. We can see that preprocessing has removed the large margins located all around the document, which was increasing the complexity of decoding for the TR+IE architecture without providing any information.

\newpage
\subsubsection{Results on the validation set for the TR+IE experiments}

Table \ref{tab:ner-table-cer} presents the Text Recognition (TR) and Information Extraction (IE) outcomes obtained on the validation set across the TR+IE experiments described in this article. The performance in information extraction mirrors that observed in the test set, with encoding number 5 continuing to exhibit superior results. Similar trends are noted in printed text recognition. 
However, we observe greater variability in the CER and WER in handwritten text recognition. We have identified that encoding methods utilizing opening and closing tags tend to underperform compared to others. This discrepancy may be attributed to the use of the F1 metric as a training stop criterion. Given that this metric is less effective for these types of encodings, it could precipitate premature or delayed cessation of training, adversely affecting text recognition performance.

\begin{table}[]
\centering
\caption{Information extraction results for different configurations.}
\label{tab:ner-table-cer}
\begin{tabular}{|c|c|c|c|c|c|}
\hline
Writing type & \begin{tabular}[c]{@{}c@{}}Entity encoding\\ (Encoding type No.)\end{tabular} & \begin{tabular}[c]{@{}c@{}}Hierarchical\\ entities\end{tabular} & \begin{tabular}[c]{@{}c@{}}CER\\ valid (\%)\end{tabular}  & \begin{tabular}[c]{@{}c@{}}WER\\ valid (\%)\end{tabular} &\begin{tabular}[c]{@{}c@{}}F1\\ valid (\%)\end{tabular} \\ \hline
Printed  & before (1)& yes   & 1.16   & 3.53 & 87.92 \\ %
Printed  & after (2)& yes &   2.39 & 4.94 &   88.25 \\ %
Printed  & after and before (3)& yes &   1.79   & 4.13 & 73.8 \\ %
Printed & after and before nested (4)& yes &  1.13 & 3.49   & 87.69 \\ %
Printed  & after (5)& no   & 1.53  & 3.9 & 92.93 \\ \hline%
Handwritten  & before (1)& yes   & 6.16   & 16.69 & 67.03 \\ %
Handwritten  & after (2)& yes &   7.16 & 17.37 &   70.6 \\ %
Handwritten  & after and before (3)& yes &   8.56   & 18.38 & 34.74 \\ %
Handwritten  & after and before nested (4)& yes &  8.04   & 18.23 & 34.91 \\ %
Handwritten  & after (5)& no &  5.63 & 15.91 & 72.65 \\ \hline
\end{tabular}%
\end{table}

\end{document}